\documentclass{article} 
\usepackage{acl2016}
\usepackage{xcolor}
\usepackage[colorlinks, linkcolor=black, urlcolor=black, citecolor=black]{hyperref}
\usepackage{url}
\usepackage{booktabs}
\usepackage{textcomp}
\usepackage{adjustbox}
\usepackage{graphicx}
\usepackage{subcaption}
\usepackage{amsmath,amssymb}
\usepackage{natbib}

\newcommand{\x}{x}
\newcommand{\z}{\vec{z}}
\newcommand\abbr[1]{\textsc{#1}}

\aclfinalcopy

\title{Generating Sentences from a Continuous Space}

\author{Samuel R. Bowman\thanks{~First two authors contributed equally. Work was done when all authors were at Google, Inc.}\\
NLP Group and Dept. of Linguistics\\
Stanford University\\
\texttt{sbowman@stanford.edu}
\And
Luke Vilnis\footnotemark[1]\\ 
CICS\\
University of Massachusetts Amherst\\
\texttt{luke@cs.umass.edu} 
\AND
Oriol Vinyals, Andrew M. Dai, Rafal Jozefowicz \& Samy Bengio\\
Google Brain\\
\texttt{\{vinyals, adai, rafalj, bengio\}@google.com} \\
}

\def\t#1{#1}
\def\b#1{\t{\textbf{#1}}}
\def\colspaceS{2.25mm}

\def\colspaceL{4.25mm}

\begin{document}

\maketitle


\begin{abstract}
The standard recurrent neural network language model (\abbr{rnnlm}) generates sentences one word at a time and does not work from an explicit global sentence representation. In this work, we introduce and study an \abbr{rnn}-based variational autoencoder generative model that incorporates distributed latent representations of entire sentences. This factorization allows it to explicitly model holistic properties of sentences such as style, topic, and high-level syntactic features. Samples from the prior over these sentence representations remarkably produce diverse and well-formed sentences through simple deterministic decoding. By examining paths through this latent space, we are able to generate coherent novel sentences that interpolate between known sentences. We present techniques for solving the difficult learning problem presented by this model, demonstrate its effectiveness in imputing missing words, explore many interesting properties of the model's latent sentence space, and present negative results on the use of the model in language modeling.
\end{abstract}


\section{Introduction}

Recurrent neural network language models \citep[\abbr{rnnlm}s,][]{mikolov2011extensions} represent the state of the art in unsupervised generative modeling for natural language sentences. In supervised settings, \abbr{rnnlm} decoders conditioned on task-specific features are the state of the art in tasks like machine translation~\citep{sutskever2014sequence, bahdanau2014neural} and image captioning \citep{vinyalscaptions,baidu_captioning,berkeley_captioning}. The \abbr{rnnlm} generates sentences word-by-word based on an evolving distributed state representation, which makes it a probabilistic model with no significant independence assumptions, and makes it capable of modeling complex distributions over sequences, including those with long-term dependencies. However, by breaking the model structure down into a series of next-step predictions, the \abbr{rnnlm} does not expose an interpretable representation of global features like topic or of high-level syntactic properties.

We propose an extension of the \abbr{rnnlm} that is designed to explicitly capture such global features in a continuous latent variable. Naively, maximum likelihood learning in such a model presents an intractable inference problem. Drawing inspiration from recent successes in modeling images \citep{gregor2015draw}, handwriting, and natural speech \citep{chung2015recurrent}, our model circumvents these difficulties using the architecture of a \emph{variational autoencoder} and takes advantage of recent advances in variational inference \citep{kingma2015auto,rezende2014stochastic} that introduce a practical training technique for powerful neural network generative models with latent variables.

\begin{table}[t]\small
\begin{center}
\begin{tabular}{l}
\toprule
\b{i went to the store to buy some groceries .}\\
{\it i store to buy some groceries .}\\
{\it i were to buy any groceries .}\\
{\it horses are to buy any groceries .}\\
{\it horses are to buy any animal .}\\
{\it horses the favorite any animal .}\\
{\it horses the favorite favorite animal .}\\
\b{horses are my favorite animal .}\\
\bottomrule
\end{tabular}
\end{center}
\caption{
\label{tab:aehomotopy}
Sentences produced by greedily decoding from points between two sentence encodings with a  conventional autoencoder. The intermediate sentences are not plausible English.
}
\end{table}

Our contributions are as follows: 
We propose a variational autoencoder architecture for text and discuss some of the obstacles to training it as well as our proposed solutions.
We find that on a standard language modeling evaluation where a global variable is not explicitly needed, this model yields similar performance to existing \abbr{rnnlm}s.
We also evaluate our model using a larger corpus on the task of imputing missing words. For this task, we introduce a novel evaluation strategy using an adversarial classifier, sidestepping the issue of intractable likelihood computations by drawing inspiration from work on non-parametric two-sample tests and adversarial training. In this setting, our model's global latent variable allows it to do well where simpler models fail. 
We finally introduce several qualitative techniques for analyzing the ability of our model to learn high level features of sentences. We find that they can produce diverse, coherent sentences through purely deterministic decoding and that they can interpolate smoothly between sentences.


\section{Background}
\subsection{Unsupervised sentence encoding}

A standard \abbr{rnn} language model predicts each word of a sentence conditioned on the previous word and an evolving hidden state.
While effective, it does not learn a vector representation of the full sentence.
In order to incorporate a continuous latent sentence representation, we first need a method to map between sentences and distributed representations that can be trained in an unsupervised setting.
While no strong generative model is available for this problem, three non-generative techniques have shown promise: sequence autoencoders, skip-thought, and paragraph vector. 

Sequence autoencoders have seen some success in pre-training sequence models for supervised downstream tasks \citep{dai2015unsup} and in generating complete documents \citep{li2015hierarchical}.  An autoencoder consists of an encoder function $\varphi_{enc}$ and a probabilistic decoder model $p(x|\z =\varphi_{enc}(x))$, and maximizes the likelihood of an example $\x$ conditioned on $\z$, the learned code for $\x$. In the case of a sequence autoencoder, both encoder and decoder are \abbr{rnn}s and examples are token sequences.

Standard autoencoders are not effective at extracting for global semantic features. In Table \ref{tab:aehomotopy}, we present the results of computing a path or \emph{homotopy} between the encodings for two sentences and decoding each intermediate code. The intermediate sentences are generally ungrammatical and do not transition smoothly from one to the other. This suggests that these models do not generally learn a smooth, interpretable feature system for sentence encoding. In addition, since these models do not incorporate a prior over $\z$, they cannot be used to assign probabilities to sentences or to sample novel sentences.

Two other models have shown promise in learning sentence encodings, but cannot be used in a generative setting: Skip-thought models \citep{kiros2015skip} are unsupervised learning models that take the same model structure as a sequence autoencoder, but generate text conditioned on a neighboring sentence from the target text, instead of on the target sentence itself. 
Finally, paragraph vector models \citep{le2014distributed} are non-recurrent sentence representation models. In a paragraph vector model, the encoding of a sentence is obtained by performing gradient-based inference on a prospective encoding vector with the goal of using it to predict the words in the sentence.

\subsection{The variational autoencoder}

The variational autoencoder \citep[\abbr{vae},][]{kingma2015auto,rezende2014stochastic} is a generative model that is based on a regularized version of the standard autoencoder. This model imposes a prior distribution on the hidden codes $\z$ which enforces a regular geometry over codes and makes it possible to draw proper samples from the model using ancestral sampling.

The \abbr{vae} modifies the autoencoder architecture by replacing the deterministic function $\varphi_{enc}$ with a learned posterior \emph{recognition model}, $q(\z|\x)$. This model parametrizes an approximate posterior distribution over $\z$ (usually a diagonal Gaussian) with a neural network conditioned on $\x$. Intuitively, the \abbr{vae} learns codes not as single points, but as soft ellipsoidal \emph{regions} in latent space, forcing the codes to fill the space rather than memorizing the training data as isolated codes.

If the \abbr{vae} were trained with a standard autoencoder's reconstruction objective, it would learn to encode its inputs deterministically by making the variances in $q(\z|\x)$ vanishingly small \citep{raiko2014techniques}. Instead, the \abbr{vae} uses an objective which encourages the model to keep its posterior distributions close to a prior $p(\z)$, generally a standard Gaussian ($\mu=\vec{0}$, $\sigma=\vec{1}$). Additionally, this objective is a valid lower bound on the true log likelihood of the data, making the \abbr{vae} a generative model. This objective takes the following form:
\begin{equation}\label{eq:vlb}
\begin{split}
 \mathcal{L}(\theta; \x) &= -\abbr{kl}(q_\theta(\z|\x)||p(\z)) \\ 
 &~~~~~ + \mathbb{E}_{q_\theta(\z|\x)}[\log p_\theta(\x|\z)] \\ 
 & \le \log p(\x)\;\;.
 \end{split}
\end{equation}
This forces the model to be able to decode plausible sentences from every point in the latent space that has a reasonable probability under the prior.

In the experiments presented below using \abbr{vae} models, we use diagonal Gaussians for the prior and posterior distributions $p(\z)$ and $q(\z|x)$, using the Gaussian reparameterization trick of \cite{kingma2015auto}. We train our models with stochastic gradient descent, and at each gradient step we estimate the reconstruction cost using a single sample from $q(\z|\x)$, but compute the \abbr{kl} divergence term of the cost function in closed form, again following~\cite{kingma2015auto}.

\section{A VAE for sentences}

\begin{figure}[t]
\begin{center}
\includegraphics[width=0.9\linewidth]{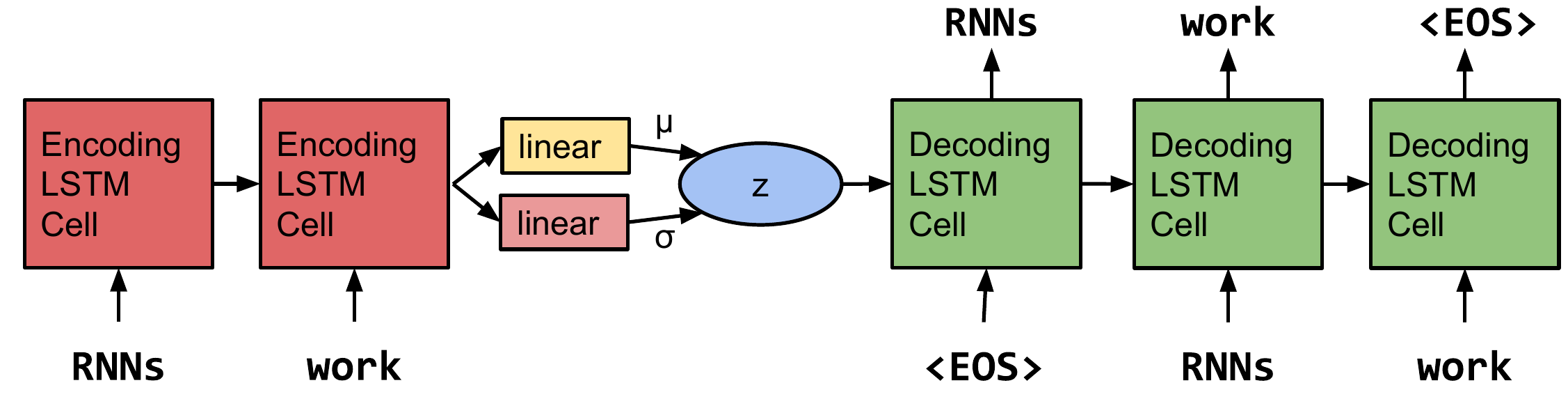}
\end{center}
\caption{\label{fig:vrae}The core structure of our variational autoencoder language model. Words are represented using a learned dictionary of embedding vectors.}
\end{figure}

We adapt the variational autoencoder to text by using single-layer \abbr{lstm} \abbr{rnn}s~\citep{hochreiter1997long} for both the encoder and the decoder, essentially forming a sequence autoencoder with the Gaussian prior acting as a regularizer on the hidden code. The decoder serves as a special \abbr{rnn} language model that is conditioned on this hidden code, and in the degenerate setting where the hidden code incorporates no useful information, this model is effectively equivalent to an \abbr{rnnlm}. The model is depicted in Figure~\ref{fig:vrae}, and is used in all of the experiments discussed below.

We explored several variations on this architecture, including concatenating the sampled $\z$ to the decoder input at every time step, using a softplus parametrization for the variance, and using deep feedforward networks between the encoder and latent variable and the decoder and latent variable. We noticed little difference in the model's performance when using any of these variations. However, when including feedforward networks between the encoder and decoder we found that it is necessary to use highway network layers \citep{srivastava2015deep}  for the model to learn. We discuss hyperparameter tuning in the appendix.

We also experimented with more sophisticated recognition models $q(\z|\x)$, including a multistep sampling model styled after \abbr{draw} \citep{gregor2015draw}, and a posterior approximation using normalizing flows \citep{rezende2015variational}. However, we were unable to reap significant gains over our plain \abbr{vae}.

While the strongest results with \abbr{vae}s to date have been on continuous domains like images, there has been some work on discrete sequences: a technique for doing this using \abbr{rnn} encoders and decoders, which shares the same high-level architecture as our model, was proposed under the name Variational Recurrent Autoencoder (\abbr{vrae}) for the modeling of music in \citet{fabius2014vrae}. While there has been other work on including continuous latent variables in \abbr{rnn}-style models for modeling speech, handwriting, and music \citep{bayer2014storn,chung2015recurrent}, these models include separate latent variables per timestep and are unsuitable for our goal of modeling global features.

In a recent paper with goals similar to ours, \citet{miao2015var} introduce an effective VAE-based document-level language model that models texts as bags of words, rather than as sequences. They mention briefly that they have to train the encoder and decoder portions of the network in alternation rather than simultaneously, possibly as a way of addressing some of the issues that we discuss in Section \ref{challenges}.

\subsection{Optimization challenges}\label{challenges}

Our model aims to learn global latent representations of sentence content. 
We can quantify the degree to which our model learns global features by looking at the variational lower bound objective \eqref{eq:vlb}. The bound breaks into two terms: the data likelihood under the posterior (expressed as cross entropy), and the \abbr{kl} divergence of the posterior from the prior. A model that encodes useful information in the latent variable $\z$ will have a non-zero \abbr{kl} divergence term and a relatively small cross entropy term. Straightforward implementations of our \abbr{vae} fail to learn this behavior: except in vanishingly rare cases, most training runs with most hyperparameters yield models that consistently set $q(\z|\x)$ equal to the prior $p(\z)$, bringing the \abbr{kl} divergence term of the cost function to zero.

When the model does this, it is essentially behaving as an \abbr{rnnlm}. Because of this, it can express arbitrary distributions over the output sentences (albeit with a potentially awkward left-to-right factorization) and can thereby achieve likelihoods that are close to optimal. Previous work on \abbr{vae}s for image modeling \citep{kingma2015auto} used a much weaker independent pixel decoder model $p(\x|\z)$, forcing the model to use the global latent variable to achieve good likelihoods. In a related result, recent approaches to image generation that use \abbr{lstm} decoders are able to do well without \abbr{vae}-style global latent variables \citep{theis2015generative}.

This problematic tendency in learning is compounded by the \abbr{lstm} decoder's sensitivity to subtle variation in the hidden states, such as that introduced by the posterior sampling process. This causes the model to initially learn to ignore $\z$ and go after low hanging fruit, explaining the data with the more easily optimized decoder. Once this has happened, the decoder ignores the encoder and little to no gradient signal passes between the two, yielding an undesirable stable equilibrium with the \abbr{kl} cost term at zero. We propose two techniques to mitigate this issue.

\paragraph{KL cost annealing}
In this simple approach to this problem, we add a variable weight to the \abbr{kl} term in the cost function at training time. At the start of training, we set that weight to zero, so that the model learns to encode as much information in $\z$ as it can. Then, as training progresses, we gradually increase this weight, forcing the model to smooth out its encodings and pack them into the prior. We increase this weight until it reaches 1, at which point the weighted cost function is equivalent to the true variational lower bound. In this setting, we do not optimize the proper lower bound on the training data likelihood during the early stages of training, but we nonetheless see improvements on the value of that bound at convergence. This can be thought of as annealing from a vanilla autoencoder to a \abbr{vae}. The rate of this increase is tuned as a hyperparameter.

\begin{figure}[t]
\centering
\includegraphics[width=0.9\linewidth]{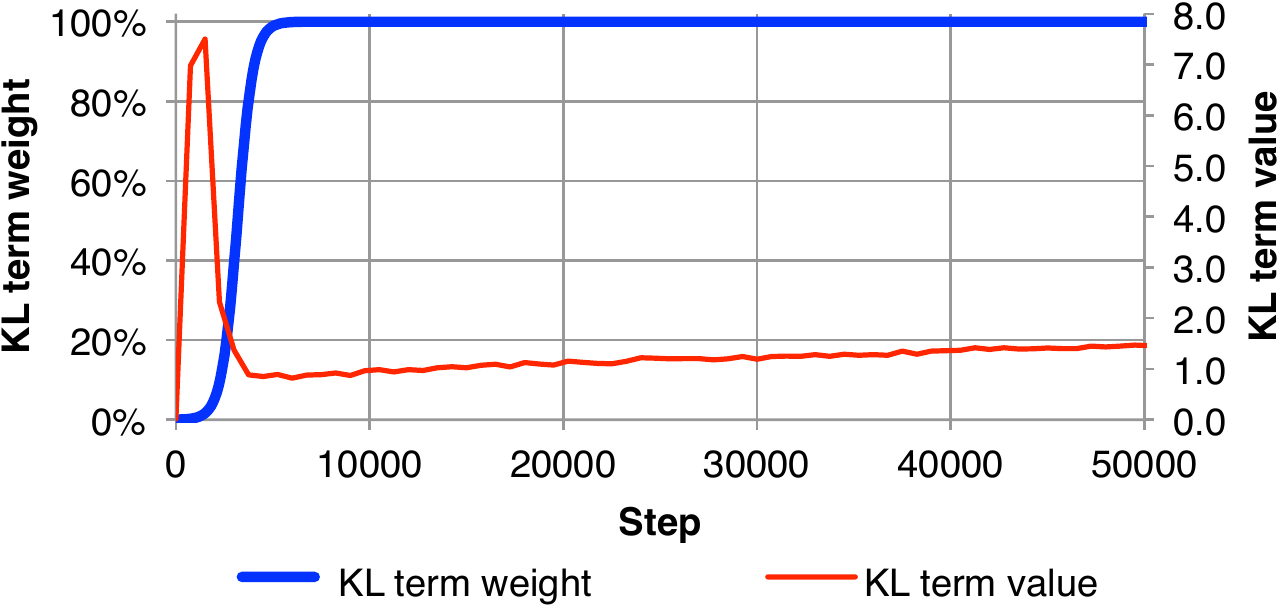}
\caption{\label{fig:klweight}The weight of the \abbr{kl} divergence term of variational lower bound according to a typical sigmoid annealing schedule plotted alongside the (unweighted) value of the \abbr{kl} divergence term for our \abbr{vae} on the Penn Treebank.}
\end{figure}

Figure~\ref{fig:klweight} shows the behavior of the \abbr{kl} cost term during the first 50k steps of training on Penn Treebank \citep{marcus1993building} language modeling with \abbr{kl} cost annealing in place. This example reflects a pattern that we observed often: \abbr{kl} spikes early in training while the model can encode information in $\z$ cheaply, then drops substantially once it begins paying the full \abbr{kl} divergence penalty, and finally slowly rises again before converging as the model learns to condense more information into $\z$.

\paragraph{Word dropout and historyless decoding}
In addition to weakening the penalty term on the encodings, we also experiment with weakening the decoder. 
As in \abbr{rnnlm}s and sequence autoencoders, during learning our decoder predicts each word conditioned on the ground-truth previous word. A natural way to weaken the decoder is to remove some or all of this conditioning information during learning. We do this by randomly replacing some fraction of the conditioned-on word tokens with the generic unknown word token \emph{\abbr{unk}}. This forces the model to rely on the latent variable $\z$ to make good predictions. This technique is a variant of word dropout \citep[][]{iyyer-EtAl:2015:ACL-IJCNLP,kumar2015ask}, applied not to a feature extractor but to a decoder. We also experimented with standard dropout \citep{srivastava2014dropout} applied to the input word embeddings in the decoder, but this did not help the model learn to use the latent variable.

This technique is parameterized by a keep rate $k\in[0,1]$. We tune this parameter both for our \abbr{vae} and for our baseline \abbr{rnnlm}. Taken to the extreme of $k=0$, the decoder sees no input, and is thus able to condition only on the number of words produced so far, yielding a model that is extremely limited in the kinds of distributions it can model without using $\z$.


\begin{table*}[ht]
\small
\begin{center}
\begin{tabular}{l@{\hskip \colspaceL}r@{~}l@{\hskip \colspaceS}r@{\hskip \colspaceS}r@{~}l@{\hskip \colspaceS}r@{\hskip \colspaceL}r@{~}l@{\hskip \colspaceS}r@{\hskip \colspaceS}r@{~}l@{\hskip \colspaceS}r}
\toprule
\b{Model} & \multicolumn{6}{c}{\b{Standard}} &  \multicolumn{6}{c}{\b{Inputless Decoder}}\\
~ & \multicolumn{2}{l}{Train \textsc{nll}} & \t{Train \textsc{ppl}} & \multicolumn{2}{l}{Test \textsc{nll}} & \t{Test \textsc{ppl}} & \multicolumn{2}{l}{Train \textsc{nll}} & \t{Train \textsc{ppl}} & \multicolumn{2}{l}{Test \textsc{nll}} & \t{Test \textsc{ppl}} \\
\midrule
\b{RNNLM} & \t{100} & \t{~--} & \t{95} & \b{100} & \t{~--} & \b{116} & \t{135} & \t{~--}  & \t{600} &\t{135} &\t{~--} & \t{$>600$}\\
\b{VAE} & \t{98} & \t{(2)} & \t{100} & \t{101} &  \t{(2)} & \t{119} & \t{120} &  \t{(15)} & \t{300} & \t{\textbf{125}} &  \t{(15)} & \b{380}\\
\bottomrule
\end{tabular}
\end{center}
\caption{
\label{tab:lmresults}
Penn Treebank language modeling results, reported as negative log likelihoods and as perplexities. Lower is better for both metrics. For the \abbr{vae}, the \abbr{kl} term of the likelihood is shown in parentheses alongside the total likelihood.
}
\end{table*}

\section{Results: Language modeling}\label{sec:lm}

In this section, we report on language modeling experiments on the Penn Treebank in an effort to discover whether the inclusion of a global latent variable is helpful for this standard task. For this reason, we restrict our \abbr{vae} hyperparameter search to those models which encode a non-trivial amount in the latent variable, as measured by the \abbr{kl} divergence term of the variational lower bound.

\paragraph{Results} We used the standard train--test split for the corpus, and report test set results in Table \ref{tab:lmresults}. The results shown reflect the training and test set performance of each model at the training step at which the model performs best on the development set. Our reported figures for the \abbr{vae} reflect the variational lower bound on the test likelihood, while for the \abbr{rnnlm}s, which can be evaluated exactly, we report the true test likelihood. This discrepancy puts the \abbr{vae} at a potential disadvantage.

In the standard setting, the \abbr{vae} performs slightly worse than the \abbr{rnnlm} baseline, though it does succeed in using the latent space to a limited extent: it has a reconstruction cost (99) better than that of the baseline \abbr{rnnlm}, but makes up for this with a \abbr{kl} divergence cost of 2. Training a \abbr{vae} in the standard setting without both word dropout and cost annealing reliably results in models with equivalent performance to the baseline \abbr{rnnlm}, and zero \abbr{kl} divergence.

To demonstrate the ability of the latent variable to encode the full content of sentences in addition to more abstract global features, we also provide numbers for an inputless decoder that does not condition on previous tokens, corresponding to a word dropout keep rate of $0$. In this regime we can see that the variational lower bound contains a significantly larger \abbr{kl} term and shows a substantial improvement over the weakened \abbr{rnnlm}, which is essentially limited to using unigram statistics in this setting. 
While it is weaker than a standard decoder, the inputless decoder has the interesting property that its sentence generating process is fully differentiable. Advances in generative models of this kind could be promising as a means of generating text while using adversarial training methods, which require differentiable generators.

Even with the techniques described in the previous section, including the inputless decoder, we were unable to train models for which the \abbr{kl} divergence term of the cost function dominates the reconstruction term. This suggests that it is still substantially easier to learn to factor the data distribution using simple local statistics, as in the \abbr{rnnlm}, such that an encoder will only learn to encode information in $\z$ when that information cannot be effectively described by these local statistics.


\begin{table*}[th]
\begin{center}\small
\begin{tabular}{lll}
\toprule
\multicolumn{3}{l}{\it but now , as they parked out front and owen stepped out of the car , he could see \underline{\hspace{0.5em}} \underline{\hspace{0.5em}} \underline{\hspace{0.5em}} \underline{\hspace{0.5em}} \underline{\hspace{0.5em}} \underline{\hspace{0.5em}}}\\
\b{True:} {\it that the transition was complete .}&
\b{RNNLM:} {\it it , '' i said .}&
\b{VAE:} {\it through the driver 's door .}\\
\midrule
\multicolumn{3}{l}{\it you kill him and his \underline{\hspace{0.5em}} \underline{\hspace{0.5em}}}\\
\b{True:} {\it men .}&
\b{RNNLM:} {\it . ''}&
\b{VAE:} {\it brother .}\\
\midrule
\multicolumn{3}{l}{\it not surprising , the mothers dont exactly see eye to eye with me \underline{\hspace{0.5em}} \underline{\hspace{0.5em}} \underline{\hspace{0.5em}} \underline{\hspace{0.5em}}}\\
\b{True:} {\it on this matter .}&
\b{RNNLM:} {\it , i said .}&
\b{VAE:} {\it , right now .}\\
\bottomrule
\end{tabular}
\end{center}
\caption{
\label{tab:imputationresults}
Examples of using beam search to impute missing words within sentences. Since we decode from right to left, note the stereotypical completions given by the \abbr{rnnlm}, compared to the \abbr{vae} completions that often use topic data and more varied vocabulary.
}
\end{table*}

\section{Results: Imputing missing words}

We claim that the our \abbr{vae}'s global sentence features make it especially well suited to the task of imputing missing words in otherwise known sentences. In this section, we present a technique for imputation and a novel evaluation strategy inspired by adversarial training. Qualitatively, we find that the \abbr{vae} yields more diverse and plausible imputations for the same amount of computation (see the examples given in Table \ref{tab:imputationresults}), but precise quantitative evaluation requires intractable likelihood computations. We sidestep this by introducing a novel evaluation strategy.

While the standard \abbr{rnnlm} is a powerful generative model, the sequential nature of likelihood computation and decoding makes it unsuitable for performing inference over unknown words given some known words (the task of \emph{imputation}).
Except in the special case where the unknown words all appear at the end of the decoding sequence, sampling from the posterior over the missing variables is intractable for all but the smallest vocabularies.
For a vocabulary of size $V$, it requires $O(V)$ runs of full \abbr{rnn} inference per step of Gibbs sampling or iterated conditional modes.
Worse, because of the directional nature of the graphical model given by an \abbr{rnnlm}, many steps of sampling could be required to propagate information between unknown variables and the known downstream variables.
The \abbr{vae}, while it suffers from the same intractability problems when sampling or computing \abbr{map} imputations, can more easily propagate information between all variables, by virtue of having a global latent variable and a tractable recognition model.

For this experiment and subsequent analysis, we train our models on the Books Corpus introduced in \cite{kiros2015skip}. 
This is a collection of text from 12k e-books, mostly fiction. 
The dataset, after pruning, contains approximately 80m sentences. We find that this much larger amount of data produces more subjectively interesting generative models than smaller standard language modeling datasets. We use a fixed word dropout rate of 75\% when training this model and all subsequent models unless otherwise specified. Our models (the \abbr{vae} and \abbr{rnnlm}) are trained as language models, decoding right-to-left to shorten the dependencies during learning for the \abbr{vae}. We use 512 hidden units.

\paragraph{Inference method} To generate imputations from the two models, we use beam search with beam size 15 for the \abbr{rnnlm} and approximate iterated conditional modes \citep{besag1986statistical} with 3 steps of a beam size 5 search for the \abbr{vae}. This allows us to compare the same amount of computation for both models. We find that breaking decoding for the \abbr{vae} into several sequential steps is necessary to propagate information among the variables.
Iterated conditional modes is a technique for finding the maximum joint assignment of a set of variables by alternately maximizing conditional distributions, and is a generalization of ``hard-\abbr{em}'' algorithms like k-means \citep{kearns1998information}.
For approximate iterated conditional modes, we first initialize the unknown words to the \emph{\abbr{unk}} token.
We then alternate assigning the latent variable to its mode from the recognition model, and performing constrained beam search to assign the unknown words.
Both of our generative models are trained to decode sentences from right-to-left, which shortens the dependencies involved in learning for the \abbr{\abbr{vae}}, and we impute the final 20\% of each sentence. This lets us demonstrate the advantages of the global latent variable in the regime where the \abbr{rnnlm} suffers the most from its inductive bias. 

\paragraph{Adversarial evaluation} Drawing inspiration from adversarial training methods for generative models as well as non-parametric two-sample tests \citep{goodfellow2014generative,li2015generative,denton2015deep,gretton2012kernel}, we evaluate the imputed sentence completions by examining their distinguishability from the true sentence endings.
While the non-differentiability of the discrete \abbr{rnn} decoder prevents us from easily applying the adversarial criterion at train time, we can define a very flexible test time evaluation by training a discriminant function to separate the generated and true sentences, which defines an \emph{adversarial error}.

We train two classifiers: a bag-of-unigrams logistic regression classifier and an \abbr{lstm} logistic regression classifier that reads the input sentence and produces a binary prediction after seeing the final \emph{\abbr{eos}} token.
We train these classifiers using early stopping on a $80/10/10$ train/dev/test split of 320k sentences, constructing a dataset of 50\% complete sentences from the corpus (positive examples) and 50\% sentences with imputed completions (negative examples). We define the \emph{adversarial error} as the gap between the ideal accuracy of the discriminator (50\%, i.e. indistinguishable samples), and the actual accuracy attained.

\begin{table}[t]
\small
\begin{center}
\begin{tabular}{lccc}
\toprule
\b{Model} & \multicolumn{2}{c}{\bf Adv. Err. (\%)} & \b{NLL}\\
 & \t{Unigram} & \abbr{lstm} & \abbr{rnnlm}\\
\midrule
\b{RNNLM (15 bm.)} & \t{28.32} & \t{38.92} & \t{46.01}\\
\b{VAE (3x5 bm.)} & \b{22.39} & \b{35.59} & \t{46.14}\\
\bottomrule
\end{tabular}
\end{center}
\caption{
\label{tab:advresults}
Results for adversarial evaluation of imputations. Unigram and \abbr{lstm} numbers are the \emph{adversarial error} (see text) and \abbr{rnnlm} numbers are the negative log-likelihood given to entire generated sentence by the \abbr{rnnlm}, a measure of sentence typicality. Lower is better on both metrics. The \abbr{vae} is able to generate imputations that are significantly more difficult to distinguish from the true sentences.
}
\end{table}

\paragraph{Results} As a consequence of this experimental setup, the \abbr{rnnlm} cannot choose anything outside of the top 15 tokens given by the \abbr{rnn}'s initial unconditional distribution $P(x_1|\text{Null})$ when producing the final token of the sentence, since it has not yet generated anything to condition on, and has a beam size of 15.
Table~\ref{tab:advresults} shows that this weakness makes the \abbr{rnnlm} produce far less diverse samples than the \abbr{vae} and suffer accordingly versus the adversarial classifier.
Additionally, we include the score given to the entire sentence with the imputed completion given a separate independently trained language model.
The likelihood results are comparable, though the \abbr{rnnlm}s favoring of generic high-probability endings such as ``he said,'' gives it a slightly lower negative log-likelihood.
Measuring the \abbr{rnnlm} likelihood of sentences themselves produced by an \abbr{rnnlm} is not a good measure of the power of the model, but demonstrates that the \abbr{rnnlm} can produce what it sees as high-quality imputations by favoring typical local statistics, even though their repetitive nature produces easy failure modes for the adversarial classifier.
Accordingly, under the adversarial evaluation our model substantially outperforms the baseline since it is able to efficiently propagate information bidirectionally through the latent variable.


\begin{table*}[ht]
\small
\begin{center}
\begin{tabular}{p{0.45\textwidth}p{0.45\textwidth}}
\toprule
\b{100\% word keep}                 &\b{75\% word keep}\\
\midrule
\textit{`` no , '' he said .}                   &\textit{`` love you , too . '' }\\
\textit{`` thank you , '' he said .}            &\textit{she put her hand on his shoulder and followed him to the door .}\\
\midrule
\b{50\% word keep}                  &\b{0\% word keep}\\
\midrule
\textit{`` maybe two or two . '' } &\textit{i i hear some of of of}\\
\textit{she laughed again , once again , once again , and thought about it for a moment in long silence .}          &\textit{i was noticed that she was holding the in in of the the in}\\
\bottomrule
\end{tabular}
\end{center}
\caption{
\label{tab:dropoutsamples}
Samples from a model trained with varying amounts of word dropout. We sample a vector from the Gaussian prior and apply greedy decoding to the result. Note that diverse samples can be achieved using a purely deterministic decoding procedure. Once we use reach a purely inputless decoder in the 0\% setting, however, the samples cease to be plausible English sentences.
}
\end{table*}

\begin{table*}
\small
\begin{center}
\begin{tabular}{l}
\toprule
\textit{he had been unable to conceal the fact that there was a logical explanation for his inability to}\\ 
\textit{alter the fact that they were supposed to be on the other side of the house .}\\
\midrule
\textit{with a variety of pots strewn scattered across the vast expanse of the high ceiling , a vase of}\\
\textit{colorful flowers adorned the tops of the rose petals littered the floor and littered the floor .}\\
\midrule
\textit{atop the circular dais perched atop the gleaming marble columns began to emerge from atop the}\\
\textit{stone dais, perched atop the dais .}\\
\bottomrule
\end{tabular}
\end{center}
 \caption{
\label{tab:bizarresamples}
Greedily decoded sentences from a model with 75\% word keep probability, sampling from lower-likelihood areas of the latent space. Note the consistent topics and vocabulary usage.}
\end{table*}

\section{Analyzing variational models}

We now turn to more qualitative analysis of the model. Since our decoder model $p(\x|\z)$ is a sophisticated \abbr{rnnlm}, simply sampling from the directed graphical model (first $p(\z)$ then $p(\x|\z)$) would not tell us much about how much of the data is being explained by each of the latent space and the decoder.
Instead, for this part of the evaluation, we sample from the Gaussian prior, but use a greedy deterministic decoder for $p(\x|\z)$, the \abbr{rnnlm} conditioned on $\z$.
This allows us to get a sense of how much of the variance in the data distribution is being captured by the distributed vector $\z$ as opposed to the decoder.
Interestingly, these results qualitatively demonstrate that large amounts of variation in generated language can be achieved by following this procedure. In the appendix, we provide some results on small text classification tasks.

\subsection{Analyzing the impact of word dropout}

\begin{figure}[t]
\begin{center}
\includegraphics[width=0.9\linewidth]{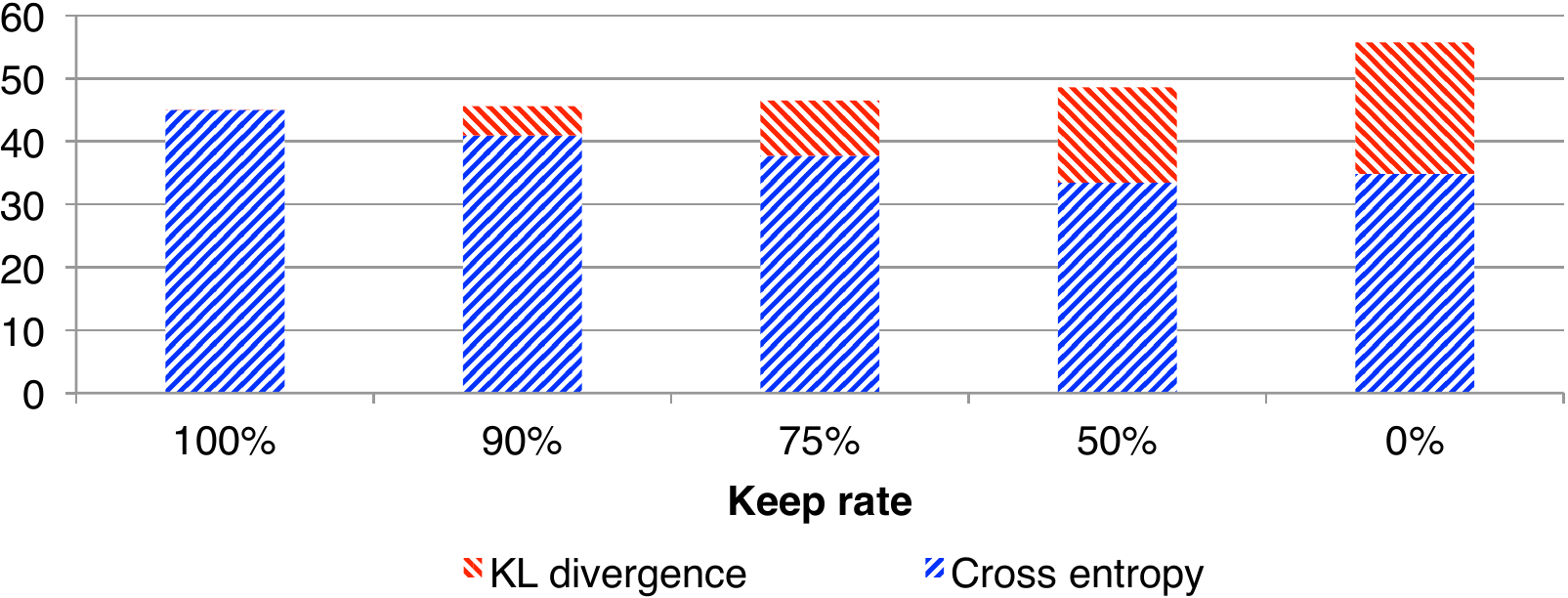}
\end{center}
\caption{\label{fig:dropoutkl}The values of the two terms of the cost function as word dropout increases.}
\end{figure}

\begin{table*}[t]
\small
\begin{center}
\begin{tabular}{llll}
\toprule
\textsc{input}  &   \b{we looked out at the setting sun .}          &\b{ i went to the kitchen .}       &\b{how are you doing ?}\\
\textsc{mean}   &   {\it they were laughing at the same time .}         &{\it  i went to the kitchen .}     &{\it what are you doing ?}\\
\textsc{samp. 1}    &   {\it ill see you in the early morning .}        &{\it i went to my apartment .} &{\it `` are you sure ?}\\
\textsc{samp. 2}    &   {\it i looked up at the blue sky .}             &{\it i looked around the room .}   &{\it  what are you doing ?}\\
\textsc{samp. 3}    &   {\it it was down on the dance floor .}          &{\it i turned back to the table .} &{\it what are you doing ?}\\
\bottomrule
\end{tabular}
\end{center}
\caption{
\label{tab:vaeposterior}
Three sentences which were used as inputs to the \abbr{vae}, presented with greedy decodes from the mean of the posterior distribution, and from three samples from that distribution.
}
\end{table*}

For this experiment, we train on the Books Corpus and test on a held out 10k sentence test set from that corpus. We find that train and test set performance are very similar. In Figure \ref{fig:dropoutkl}, we examine the impact of word dropout on the variational lower bound, broken down into \abbr{kl} divergence and cross entropy components. We drop out words with the specified keep rate at training time, but supply all words as inputs at test time except in the $0\%$ setting.

We do not re-tune the hyperparameters for each run, which results in the model with no dropout encoding very little information in $\z$ (i.e., the \abbr{kl} component is small). 
We can see that as we lower the keep rate for word dropout, the amount of information stored in the latent variable increases, and the overall likelihood of the model degrades somewhat. Results from the Section~\ref{sec:lm} indicate that a model with no latent variable would degrade in performance significantly more in the presence of heavy word dropout. 

We also qualitatively evaluate samples, to demonstrate that the increased \abbr{kl} allows meaningful sentences to be generated purely from continuous sampling. Since our decoder model $p(\x|\z)$ is a sophisticated \abbr{rnnlm}, simply sampling from the directed graphical model (first $p(\z)$ then $p(\x|\z)$) would not tell us about how much of the data is being explained by the learned vector vs. the language model.
Instead, for this part of the qualitative evaluation, we sample from the Gaussian prior, but use a greedy deterministic decoder for $\x$, taking each token $x_t=\text{argmax}_{x_t} p(x_t|\x_{0,...,t-1}, \z)$. This allows us to get a sense of how much of the variance in the data distribution is being captured by the distributed vector $\z$ as opposed to by local language model dependencies. 

These results, shown in Table \ref{tab:dropoutsamples}, qualitatively demonstrate that large amounts of variation in generated language can be achieved by following this procedure. At the low end, where very little of the variance is explained by $\z$, we see that greedy decoding applied to a Gaussian sample does not produce diverse sentences. As we increase the amount of word dropout and force $\z$ to encode more information, we see the sentences become more varied, but past a certain point they begin to repeat words or show other signs of ungrammaticality.
Even in the case of a fully dropped-out decoder, the model is able to capture higher-order statistics not present in the unigram distribution.

Additionally, in Table \ref{tab:bizarresamples} we examine the effect of using lower-probability samples from the latent Gaussian space for a model with a 75\% word keep rate.
We find lower-probability samples by applying an approximately volume-preserving transformation to the Gaussian samples that stretches some eigenspaces by up to a factor of 4.
This has the effect of creating samples that are not too improbable under the prior, but still reach into the tails of the distribution.
We use a random linear transformation, with matrix elements drawn from a uniform distribution from $[ -c,c ]$, with $c$ chosen to give the desired properties ($0.1$ in our experiments).
Here we see that the sentences are far less typical, but for the most part are grammatical and maintain a clear topic, indicating that the latent variable is capturing a rich variety of global features even for rare sentences.

\subsection{Sampling from the posterior}

In addition to generating unconditional samples, we can also examine the sentences decoded from the posterior vectors $p(z|x)$ for various sentences $x$. 
Because the model is regularized to produce distributions rather than deterministic codes, it does not exactly memorize and round-trip the input.
Instead, we can see what the model considers to be similar sentences by examining the posterior samples in Table~\ref{tab:vaeposterior}.
The codes appear to capture information about the number of tokens and parts of speech for each token, as well as topic information.
As the sentences get longer, the fidelity of the round-tripped sentences decreases.

\begin{table}[t]
\small
\begin{center}
\begin{tabular}{l}
\toprule
\b{`` i want to talk to you . ''}\\
{\it ``i want to be with you . ''}\\
{\it ``i do n't want to be with you . ''}\\
{\it i do n't want to be with you .}\\
\b{she did n't want to be with him .}\\
\midrule
\b{he was silent for a long moment .}\\
{\it he was silent for a moment .}\\
{\it it was quiet for a moment .}\\
{\it it was dark and cold .}\\
{\it there was a pause .}\\
\b{it was my turn .}\\
\bottomrule
\end{tabular}
\end{center}
\caption{
\label{tab:vaehomotopy}
Paths between pairs of random points in \abbr{vae} space: Note that intermediate sentences are grammatical, and that topic and syntactic structure are usually locally consistent.}
\end{table}

\subsection{Homotopies}\label{ssec:homotopies}

The use of a variational autoencoder allows us to generate sentences using greedy decoding on continuous samples from the space of codes.
Additionally, the volume-filling and smooth nature of the code space allows us to examine for the first time a concept of \emph{homotopy} (linear interpolation) between sentences.
In this context, a homotopy between two codes $\z_1$ and $\z_2$ is the set of points on the line between them, inclusive, $\z(t)=\z_1*(1-t)+\z_2*t$ for $t \in [0, 1]$.
Similarly, the homotopy between two sentences decoded (greedily) from codes $\z_1$ and $\z_2$ is the set of sentences decoded from the codes on the line.
Examining these homotopies allows us to get a sense of what neighborhoods in code space look like -- how the autoencoder organizes information and what it regards as a continuous deformation between two sentences.

While a standard non-variational \abbr{rnnlm} does not have a way to perform these homotopies, a vanilla sequence autoencoder can do so. As mentioned earlier in the paper, if we examine the homotopies created by the sequence autoencoder in Table \ref{tab:aehomotopy}, though, we can see that the transition between sentences is sharp, and results in ungrammatical intermediate sentences.
This gives evidence for our intuition that the \abbr{vae} learns representations that are smooth and ``fill up'' the space.

In Table~\ref{tab:vaehomotopy} (and in additional tables in the appendix) we can see that the codes mostly contain syntactic information, such as the number of words and the parts of speech of tokens, and that all intermediate sentences are grammatical. Some topic information also remains consistent in neighborhoods along the path. Additionally, sentences with similar syntax and topic but flipped sentiment valence, e.g. ``the pain was unbearable'' vs. ``the thought made me smile'', can have similar embeddings, a phenomenon which has been observed with single-word embeddings (for example the vectors for ``bad'' and ``good'' are often very similar due to their similar distributional characteristics).

\section{Conclusion}

This paper introduces the use of a variational autoencoder for natural language sentences. We present novel techniques that allow us to train our model successfully, and find that it can effectively impute missing words. We analyze the latent space learned by our model, and find that it is able to generate coherent and diverse sentences through purely continuous sampling and provides interpretable homotopies that smoothly interpolate between sentences.

We hope in future work to investigate factorization of the latent variable into separate style and content components, to generate sentences conditioned on extrinsic features, to learn sentence embeddings in a semi-supervised fashion for language understanding tasks like textual entailment, and to go beyond adversarial evaluation to a fully adversarial training objective.



\bibliography{iclr_variational_sentences}
\bibliographystyle{acl_natbib}


\section*{Text classification}

\begin{table}[t]
\small
\begin{center}
\begin{tabular}{lccc}
\toprule
\b{Method} & \b{Accuracy} & \b{F1}\\
\midrule
Feats & 73.2 & --\\
\abbr{rae}+\abbr{dp} & 72.6 & --\\
\abbr{rae}+feats & 74.2 & --\\
\abbr{rae}+\abbr{dp}+feats & 76.8 & 83.6\\
\midrule
\abbr{st} & 73.0 & 81.9\\
Bi-\abbr{st} & 71.2 & 81.2\\
Combine-\abbr{st} & 73.0 & 82.0\\
\midrule
\abbr{vae} & 72.9 & 81.4\\
\abbr{vae}+feats & 75.0 & 82.4\\
\abbr{vae}+combine-\abbr{st} & 74.8 & 82.3\\
Feats+combine-\abbr{st} & 75.8 & 83.0\\
\abbr{vae}+combine-\abbr{st}+feats & \b{76.9} & \b{83.8}\\
\bottomrule
\end{tabular}
\end{center}
\caption{
\label{tab:pararesults}
Results for the \abbr{msr} Paraphrase Corpus.
}
\end{table}

In order to further examine the the structure of the representations discovered by the \abbr{vae}, we conduct classification experiments on paraphrase detection and question type classification. We train a \abbr{vae} with a hidden state size of 1200 hidden units on the Books Corpus, and use the posterior mean of the model as the extracted sentence vector. We train classifiers on these means using the same experimental protocol as \cite{kiros2015skip}.

\paragraph{Paraphrase detection} For the task of paraphrase detection, we use the Microsoft Research Paraphrase Corpus \citep{dolan2004unsupervised}. We compute features from the sentence vectors of sentence pairs in the same way as \cite{kiros2015skip}, concatenating the elementwise products and the absolute value of the elementwise differences of the two vectors. We train an $\ell_2$-regularized logistic regression classifier and tune the regularization strength using cross-validation.

We present results in Table \ref{tab:pararesults} and compare to several previous models for this task. \emph{Feats} is the lexicalized baseline from \cite{socher2011dynamic}. \emph{\abbr{rae}} uses the recursive autoencoder from that work, and \emph{\abbr{dp}} adds their dynamic pooling step to calculate pairwise features. \emph{\abbr{st}} uses features from the unidirectional skip-thought model, \emph{bi-\abbr{st}} uses bidirectional skip-thought, and \emph{combine-\abbr{st}} uses the concatenation of those features. We also experimented with concatenating lexical features and the two types of distributed features.

We found that our features performed slightly worse than skip-thought features by themselves and slightly better than recursive autoencoder features, and were complementary and yielded strong performance when simply concatenated with the skip-thought features.

\paragraph{Question classification} We also conduct experiments on the TREC Question Classification dataset of \cite{li2002learning}. Following \cite{kiros2015skip}, we train an $\ell_2$-regularized softmax classifier with 10-fold cross-validation to set the regularization. Note that using a linear classifier like this one may disadvantage our representations here, since the Gaussian distribution over hidden codes in a \abbr{vae} is likely to discourage linear separability.

\begin{table}[t]
\small
\begin{center}
\begin{tabular}{lccc}
\toprule
\b{Method} & \b{Accuracy}\\
\midrule
\abbr{st} & 91.4\\
Bi-\abbr{st}  & 89.4\\
Combine-\abbr{st}  & \b{92.2}\\
AE  & 84.2\\
\abbr{vae}  & 87.0\\
\abbr{cbow}  & 87.3\\
\abbr{vae}, combine-\abbr{st} & 92.0\\
\midrule
\abbr{rnn}  & 90.2\\
\abbr{cnn}  & \b{93.6}\\
\bottomrule
\end{tabular}
\end{center}
\caption{
\label{tab:trecresults}
Results for TREC Question Classification.}
\end{table}

We present results in Table \ref{tab:trecresults}. Here, \abbr{ae} is a plain sequence autoencoder. We compare with results from a bag of word vectors \citep[\abbr{cbow},][]{zhao2015self} and skip-thought (\abbr{st}). We also compare with an \abbr{rnn} classifier \citep{zhao2015self} and a \abbr{cnn} classifier \citep{kim2014convolutional} both of which, unlike our model, are optimized end-to-end. We were not able to make the \abbr{vae} codes perform better than \abbr{cbow} in this case, but they did outperform features from the sequence autoencoder. Skip-thought performed quite well, possibly because the skip-thought training objective of next sentence prediction is well aligned to this task: it essentially trains the model to generate sentences that address implicit open questions from the narrative of the book. Combining the two representations did not give any additional performance gain over the base skip-thought model.

\section*{Hyperparameter tuning}

\begin{table*}[ht]
\small
\begin{center}
\begin{tabular}{lrrrr}
\toprule
~   &   \multicolumn{2}{c}{\b{Standard}}    & \multicolumn{2}{c}{\b{Inputless Decoder}}\\
~   &   \abbr{rnnlm} & \abbr{vae} & \abbr{rnnlm} & \abbr{vae}\\\midrule
Embedding dim. &    464 &353&   305&    499\\
\abbr{lstm} state dim.   &337&  191&    68  &350\\
$z$ dim.    &--&    13  &--&    111\\
Word dropout keep rate &    0.66&   0.62&   --& --\\
\bottomrule
\end{tabular}
\end{center}
\caption{
\label{tab:lmparams}
Automatically selected hyperparameter values used for the models used in the Penn Treebank language modeling experiments.
}
\end{table*}

We extensively tune the hyperparameters of each model using an automatic Bayesian hyperparameter tuning algorithm \citep[based on][]{snoek2012practical} over development set data. We run the model with each set of hyperpameters for 10 hours, operating 12 experiments in parallel, and choose the best set of hyperparameters after 200 runs. Results for our language modeling experiments are reported in Table~\ref{tab:lmparams} on the next page.

\begin{table*}[h]
\small ~\\
\begin{center}
\adjustbox{valign=t}{
\begin{minipage}[t]{0.45\textwidth}\small\centering
\begin{tabular}{l}
\toprule
\b{amazing , is n't it ?}\\
{\it so , what is it ?}\\
{\it it hurts , isnt it ?}\\
{\it why would you do that ?}\\
{\it `` you can do it .}\\
{\it `` i can do it .}\\
{\it i ca n't do it .}\\
{\it `` i can do it .}\\
{\it `` do n't do it .}\\
{\it `` i can do it .}\\
\b{i could n't do it .}\\
\bottomrule
\end{tabular}\vspace{2em}
\begin{tabular}{l}
\toprule
\b{no .}\\
{\it  he said .}\\
{\it  `` no , '' he said .}\\
{\it  `` no , '' i said .}\\
{\it  `` i know , '' she said .}\\
{\it  `` thank you , '' she said .}\\
{\it  `` come with me , '' she said .}\\
{\it  `` talk to me , '' she said .}\\
\b{  `` do n't worry about it , '' she said .}\\
\bottomrule
\end{tabular}\vspace{2em}
\begin{tabular}{l}
\toprule
\b{i dont like it , he said . }\\
{\it i waited for what had happened . }\\
{\it it was almost thirty years ago . }\\
{\it it was over thirty years ago . }\\
{\it that was six years ago . }\\
{\it he had died two years ago . }\\
{\it ten , thirty years ago . }\\
{\it `` it 's all right here . }\\
{\it `` everything is all right here . }\\
{\it `` it 's all right here . }\\
{\it it 's all right here . }\\
{\it we are all right here . }\\
\b{come here in five minutes . }\\
\bottomrule
\end{tabular}
\end{minipage}\qquad
}\adjustbox{valign=t}{
\begin{minipage}[t]{0.45\textwidth}\small\centering
\begin{tabular}{l}
\toprule
\b{this was the only way .  }\\
{\it it was the only way .  }\\
{\it it was her turn to blink .  }\\
{\it it was hard to tell .  }\\
{\it it was time to move on .  }\\
{\it he had to do it again .  }\\
{\it they all looked at each other .  }\\
{\it they all turned to look back .  }\\
{\it they both turned to face him .  }\\
\b{they both turned and walked away .  }\\
\bottomrule
\end{tabular}\vspace{2em}
\begin{tabular}{l}
\toprule
\b{there is no one else in the world .  }\\
{\it there is no one else in sight .  }\\
{\it they were the only ones who mattered .  }\\
{\it they were the only ones left .  }\\
{\it he had to be with me .  }\\
{\it she had to be with him .  }\\
{\it i had to do this .  }\\
{\it i wanted to kill him .  }\\
{\it i started to cry .  }\\
\b{i turned to him .  }\\
\bottomrule
\end{tabular}\vspace{2em}
\begin{tabular}{l}
\toprule
\b{im fine . }\\
{\it youre right . }\\
{\it `` all right . }\\
{\it you 're right . }\\
{\it okay , fine . }\\
{\it `` okay , fine . }\\
{\it yes , right here . }\\
{\it no , not right now . }\\
{\it `` no , not right now . }\\
{\it `` talk to me right now . }\\
{\it please talk to me right now . }\\
{\it i 'll talk to you right now . }\\
{\it `` i 'll talk to you right now . }\\
{\it `` you need to talk to me now . }\\
\b{`` but you need to talk to me now . }\\
\bottomrule
\end{tabular}
\end{minipage}}
\end{center}
\caption{
\label{tab:vaehomotopy2}
Selected homotopies between pairs of random points in the latent \abbr{vae} space.}
\end{table*}

\section*{Additional homotopies}

Table \ref{tab:vaehomotopy2}, on the next page, shows additional homotopies from our model. We observe that intermediate sentences are almost always grammatical, and often contain consistent topic, vocabulary and syntactic information in local neighborhoods as they interpolate between the endpoint sentences. Because the model is trained on fiction, including romance novels, the topics are often rather dramatic.

\end{document}